\pdfoutput=1
\documentclass[11pt]{article}

\usepackage[final]{acl}
\usepackage{times}
\usepackage{latexsym}
\usepackage[T1]{fontenc}
\usepackage[utf8]{inputenc}
\usepackage{microtype}
\usepackage{inconsolata}
\usepackage{graphicx}
\usepackage{subcaption}
\usepackage{booktabs}
\usepackage{hyperref}
\usepackage{amsmath}
\usepackage{amssymb}
\usepackage{mathtools}
\usepackage{amsthm}
\usepackage{amsfonts}
\usepackage{multirow}
\usepackage{bbm}
\usepackage{soul}
\usepackage{xcolor}
\usepackage[capitalize,noabbrev]{cleveref}
\usepackage{array}

\newcommand{\unigr}{\mbox{\sc{Magnet}}}
\newcommand{\unigrb}{\mbox{\sc{\textbf{Magnet}}}}
\newcommand{\x}{\mathbf{x}}
\newcommand{\y}{\mathbf{y}}
\newcommand{\h}{\mathbf{h}}
\newcommand{\queries}{\mathbf{Q}}
\newcommand{\keys}{\mathbf{K}}
\newcommand{\values}{\mathbf{V}}
\newcommand{\mask}{\mathbf{M}}
\newcommand{\boldheader}[1]{\noindent\textbf{#1.}\hspace{0.12cm}}
\newcommand{\highlight}[1]{\hl{\textit{#1}}}
\newcommand{\blank}{\highlight{\textunderscore\textunderscore\textunderscore\textunderscore\textunderscore}\;}
\newcolumntype{C}[1]{>{\centering\arraybackslash}p{#1}}

\title{\unigrb{}: Augmenting Generative Decoders with Representation Learning and Infilling Capabilities}

\author{
    \fontsize{10pt}{11pt}\selectfont
    \textbf{Savya Khosla}$^{12}$\thanks{Work done during internship at Adobe Research. Correspondence to \texttt{savyak2@illinois.edu}.}, \textbf{Aditi Tiwari}$^{2}$, \textbf{Kushal Kafle}$^{1}$, \textbf{Simon Jenni}$^{1}$, \textbf{Handong Zhao}$^{1}$, \textbf{John Collomosse}$^{1}$, \textbf{Jing Shi}$^{1}$ \\
    \vspace{-0.12in} \\
    \fontsize{10pt}{11pt}\selectfont
    $^1$Adobe Research, $^2$University of Illinois Urbana-Champaign
}

\begin{document}
\maketitle

\begin{abstract}
While originally designed for unidirectional generative modeling, decoder-only large language models (LLMs) are increasingly being adapted for bidirectional modeling. However, unidirectional and bidirectional models are typically trained separately with distinct objectives (generation and representation learning). This separation overlooks the opportunity for developing a more versatile language model and for these objectives to complement each other. In this work, we propose \unigr{}, a method for adapting decoder-only LLMs to generate robust representations and infill missing text spans. \unigr{} employs three self-supervised training objectives and introduces an attention mechanism that combines bidirectional and causal attention, enabling unified training across all objectives. Our results demonstrate that LLMs adapted with \unigr{} (1) surpass strong text encoders on token-level and sentence-level representation learning tasks, (2) generate contextually appropriate text infills by leveraging past and future contexts, (3) perform open-ended text generation without excessive repetition of words or phrases, and (4) preserve the knowledge and reasoning capability gained by the LLM during pretraining.
\end{abstract}
\section{Introduction}
\label{sec:introduction}

\begin{figure*}[t]
    \centering
    \includegraphics[width=0.92\textwidth]{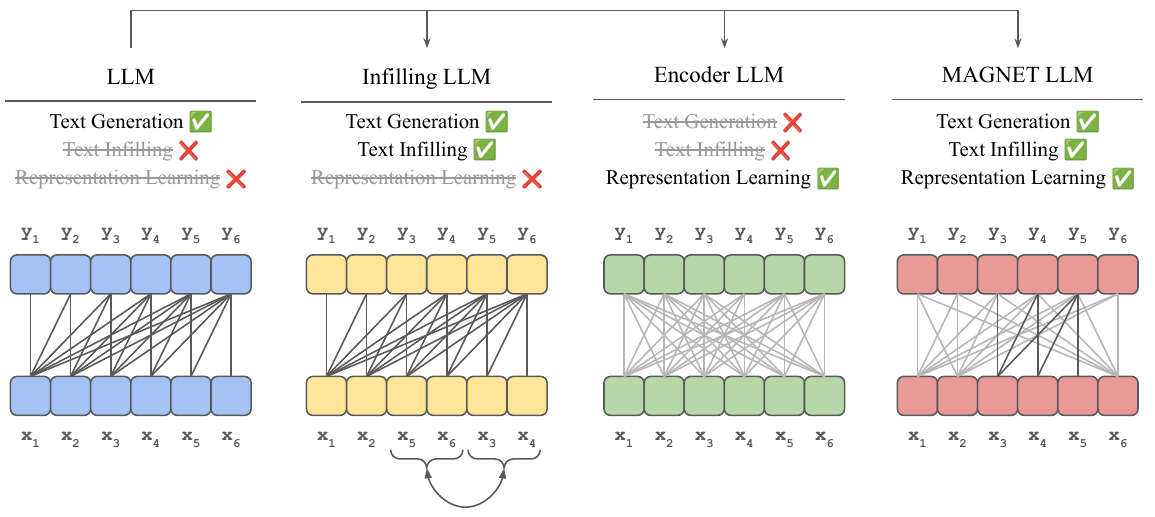}
    \vskip -0.15in
    \caption{Traditionally, LLMs are trained for text generation using unidirectional attention between the input $x$ and output $y$ (depicted by black lines), whereas text encoders are trained for representation learning using bidirectional attention (depicted by gray lines). \unigr{} adapts the attention mechanism of LLMs to combine both unidirectional and bidirectional attention, enhancing them with representation learning and infilling capabilities, while retaining their core generative functions.}
    \label{fig:teaser}
\end{figure*}

Decoder-only LLMs have gained popularity in NLP due to their efficient training and scalability. However, their reliance on causal attention restricts their effectiveness in tasks that require understanding of bidirectional context. This limitation is particularly evident in (1) representation learning tasks such as sentiment analysis and named entity recognition, where understanding the full context of sentences or words is crucial, and (2) text infilling, where filling in missing spans must ensure coherence with the surrounding text. 

Some recent efforts \citep{BehnamGhader2024LLM2VecLL, Li2023BeLLMBD, Li2023LabelSL, Dukic2024LookingRI, Du2021GLMGL, Donahue2020EnablingLM} have sought to adapt decoder-only LLMs for representation learning and text infilling. However, as shown in Figure~\ref{fig:teaser}, methods that enhance LLMs for text infilling fail to make them effective text encoders, while methods focused on representation learning diminish their generative capabilities.

In this work, we introduce \unigr{} (\underline{M}odified \underline{A}ttention for \underline{G}eneration a\underline{n}d \underline{E}ncoding of \underline{T}ext), a method for adapting decoder-only LLMs into more versatile language models. \unigr{} enables an LLM to (1) generate robust sentence-level and token-level representations, (2) infill missing text spans while maintaining coherence with bidirectional context, (3) perform open-ended text generation without excessive repetition, and (4) preserve the knowledge acquired during pretraining. In essence, \unigr{} equips an LLM with representation learning and infilling capabilities while preserving its generative strengths. To achieve this, we use three self-supervised training objectives: (1) a \textit{masked modeling objective} to learn token-level representations, (2) a \textit{contrastive objective} to learn sentence-level representations, and (3) a \textit{missing-span generation objective} to infill text and retain generative capabilities. To facilitate simultaneous training across all these objectives, we deploy a specially crafted attention mask that combines bidirectional and causal attention.

Without any model-specific design, we apply \unigr{} to Llama-2-7B \citep{Touvron2023Llama2O}. We demonstrate that the proposed method requires simple modification and fine-tuning of an off-the-shelf LLM to augment it with representation learning and infilling capabilities. Our results show that \unigr{}-adapted Llama-2-7B outperforms other methods that adapt the same model for token-level and sentence-level representation learning tasks\footnote{It is to be noted that while these other methods adapt the model exclusively for representation learning, \unigr{} incorporates additional objectives, making the LLM more versatile and showcasing the advantages of unified training.}. We also show that \unigr{} improves the infilling capability of the LLM by enabling it to consider the bidirectional context. Further, we analyze the repetition problem in text generated by models that are trained or fine-tuned to encode text and demonstrate that \unigr{}-adapted models are significantly better at open-ended text generation than other text encoders. Lastly, we show that \unigr{} preserves the knowledge and reasoning capabilities acquired by the LLM during pretraining.
\section{Related Works}
\label{sec:related}

\boldheader{Representation Learning}
Text representation learning focuses on understanding contextual relationships within sentences. Traditionally, encoder models dominated this field due to their bidirectional context modeling, using masked language modeling for token-level representations \citep{Devlin2019BERTPO, Liu2019RoBERTaAR, He2020DeBERTaDB, clark2020electra, He2021DeBERTaV3ID} and special tokens with similarity-based optimization for sentence-level understanding \citep{Gunel2020SupervisedCL,Reimers2019SentenceBERTSE, Wu2020CLEARCL, Carlsson2021SemanticRW, Gao2021SimCSESC, Wei2020OnLU}. Recent work has explored adapting decoder-only LLMs for text encoding through various methods, including introducing special tokens to the model's vocabulary~\citep{Zhang2024OneGenEO}, using last-token or mean-pooled representations \citep{Neelakantan2022TextAC, Wang2023ImprovingTE}, or fine-tuning with masked modeling \citep{BehnamGhader2024LLM2VecLL} or label supervision \citep{Li2023LabelSL, Dukic2024LookingRI}. While some approaches modify the decoder's causal attention to be bidirectional \citep{BehnamGhader2024LLM2VecLL, Muennighoff2024GenerativeRI, Li2023BeLLMBD, Dukic2024LookingRI, Man2024ULLMEAU}, this often compromises the model's text generation abilities. In contrast, \unigr{} employs a hybrid attention mechanism that combines causal and bidirectional attention, enabling both robust representation learning and preserved generation capabilities.

\boldheader{Text Infilling}
Text infilling requires considering both left and right context when generating text in the middle of a sequence. Encoder-decoder models \citep{Raffel2019ExploringTL, Lewis2019BARTDS, Kalinsky2023SimpleAE} can handle this task by encoding available context and decoding infilled text. Other approaches have extended masked language modeling to perform span infilling \citep{Joshi2019SpanBERTIP, Shen2023FiLMFL, Shen2020BlankLM}. Decoder-only models have also been adapted for infilling through various strategies: training models to directly fill marked blanks \citep{Donahue2020EnablingLM, Du2021GLMGL}, rearranging training examples to align with infilling objectives \citep{Bavarian2022EfficientTO, Yang2019XLNetGA, Aghajanyan2022CM3AC, Fried2022InCoderAG}, or using dual generation from both ends of a sentence until convergence \citep{Nguyen2023MeetIT, Serdyuk2017TwinNM}. However, while these approaches successfully enhance LLMs with infilling capabilities, none have attempted to simultaneously equip them with both infilling and representation learning abilities, as done by \unigr{}.

\boldheader{Unifying Text Understanding and Generation}
Prior works on unifying natural language understanding and generation within a single framework usually focus on proposing pretraining objectives and task formulations. These approaches typically extend traditional masked language modeling, with innovations like permutation-based objectives for bidirectional context modeling \citep{Yang2019XLNetGA}, autoregressive blank infilling \citep{Du2021GLMGL}, multi-directional attention masks \citep{Dong2019UnifiedLM}, and sequence-to-sequence pretraining \citep{Song2019MASSMS, Raffel2019ExploringTL}. However, these approaches require pretraining new networks from scratch, despite decoder-only models demonstrating exceptional scalability and effectiveness. Instead of starting from scratch, we propose a parameter-efficient method that builds upon the rich representations already learned by existing large language models, transforming them into a unified framework for representation learning, text infilling, and text generation.

\section{Method}
\label{sec:method}
\begin{figure}[t]
    \centering
    \begin{subfigure}[b]{0.2\textwidth}
        \centering
        \includegraphics[width=\textwidth]{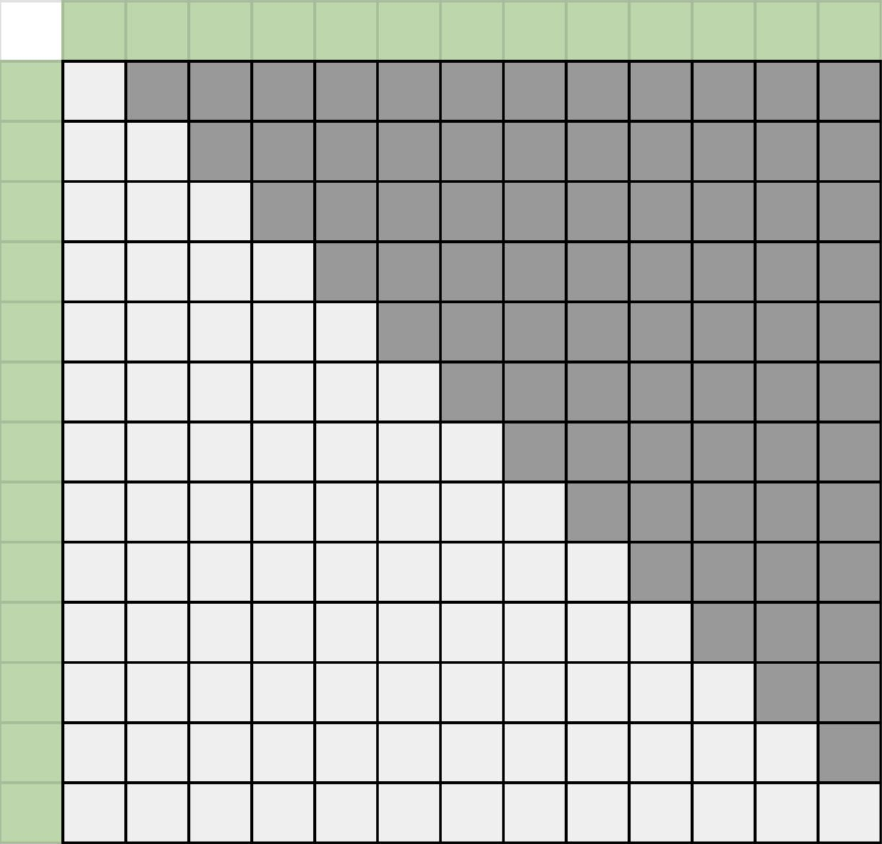}
        \caption{Causal mask}
        \label{fig:causal-mask}
    \end{subfigure}
    \quad\quad
    \begin{subfigure}[b]{0.2\textwidth}
        \centering
        \includegraphics[width=\textwidth]{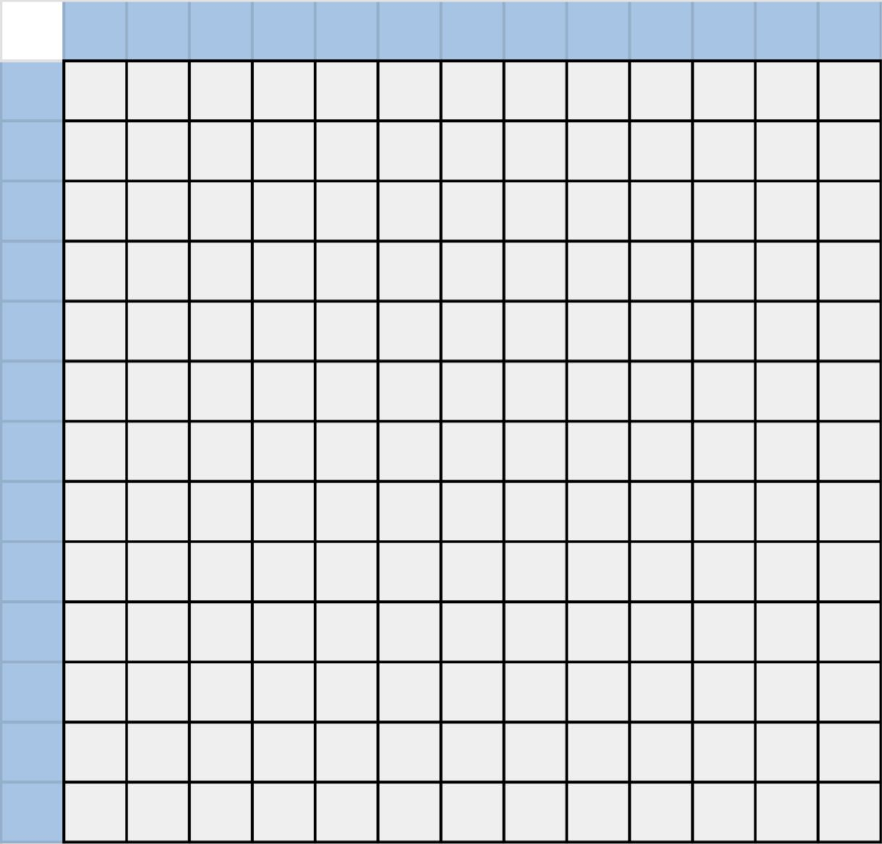}
        \caption{Bidirectional mask}
        \label{fig:bidirectional-mask}
    \end{subfigure}
    \vskip 0.1in
    \begin{subfigure}[b]{0.2\textwidth}
        \centering
        \includegraphics[width=\textwidth]{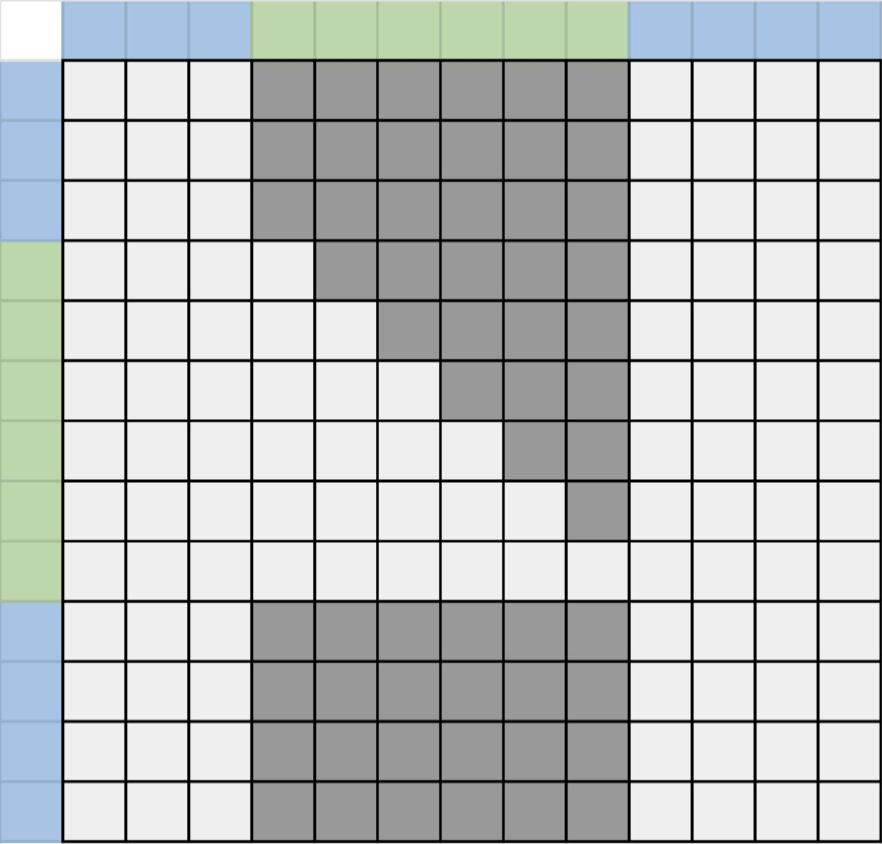}
        \caption{Mask with 1 span}
        \label{fig:unigr-mask-1}
    \end{subfigure}
    \quad\quad
    \begin{subfigure}[b]{0.2\textwidth}
        \centering
        \includegraphics[width=\textwidth]{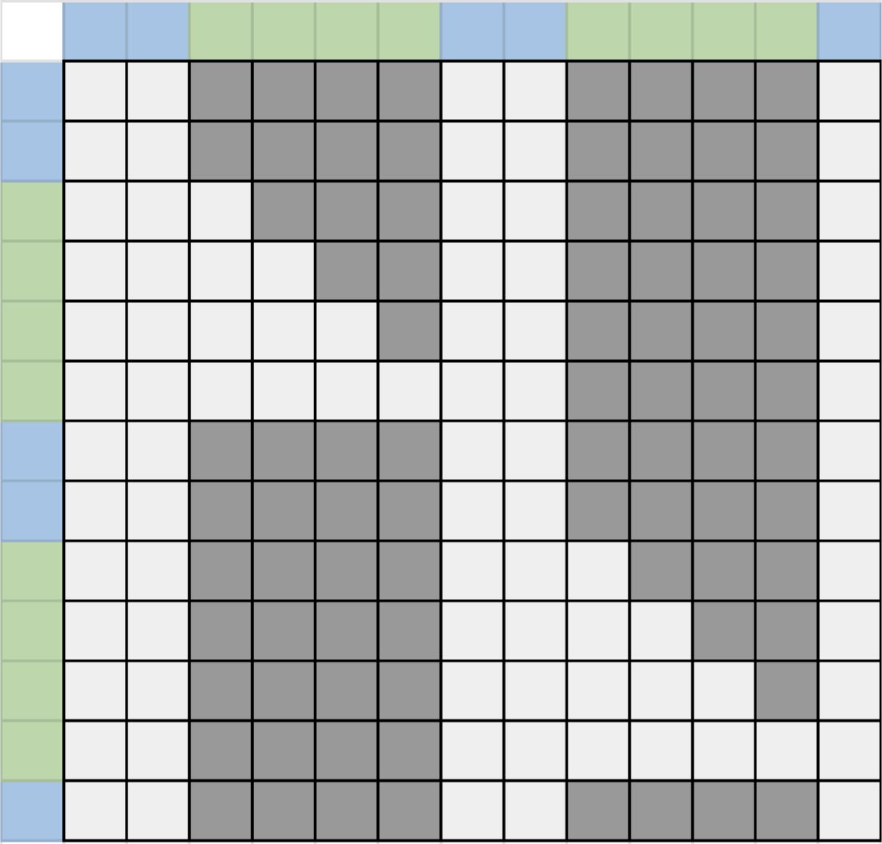}
        \caption{Mask with 2 spans}
        \label{fig:unigr-mask-2}
    \end{subfigure}
    \vskip -0.05in
    \caption{Attention masks for different types of attention mechanisms. The rows of the matrices correspond to the query tokens and the columns correspond to the key tokens. Light gray cells indicate 0, dark gray cells represent $-\infty$, green marks span token positions, and blue marks context token positions. Each context token attends to every other context token, and each span token attends to all context tokens and the preceding span tokens in the same span.}
    \label{fig:modified-attention}
    \vskip -0.15in
\end{figure}

Decoder-only models process input sequences through successive blocks of multi-head self-attention, feed-forward networks, and layer normalization. The self-attention mechanism converts the input $\x{} \in \mathbb{R}^{l \times d}$ into queries $\queries{}$, keys $\keys{}$, and values $\values{}$ using linear projections, and computes attention using the formula:
$$
    \texttt{Attn}_i(\queries{}, \keys{}, \values{}) = \texttt{softmax}\left(\frac{\queries{} \keys{}^\text{T} + \mask{}}{\sqrt{d_k}}\right) \values{}
$$
where $\texttt{Attn}_i$ is the $i^{th}$ head of the multi-head self-attention, $d_k$ represents the dimensionality of the keys/queries, and $\mask{}$ represents the causal mask. This causal mask $\mask{}$ for an autoregressive LLM is a $l \times l$ strictly upper triangular matrix, as shown in Figure~\ref{fig:causal-mask}, and it ensures that each token can only attend to itself and tokens that precede it.

\unigr{} updates the causal attention mechanism of an LLM to incorporate elements of bidirectionality and thereafter fine-tunes the model using self-supervised objectives. We look at the modifications to the attention mechanism in Section~\ref{sec:modified-attention} and the training objectives in Section~\ref{sec:training-objectives}.

\begin{figure*}[t]
    \centering
    \includegraphics[width=0.95\textwidth]{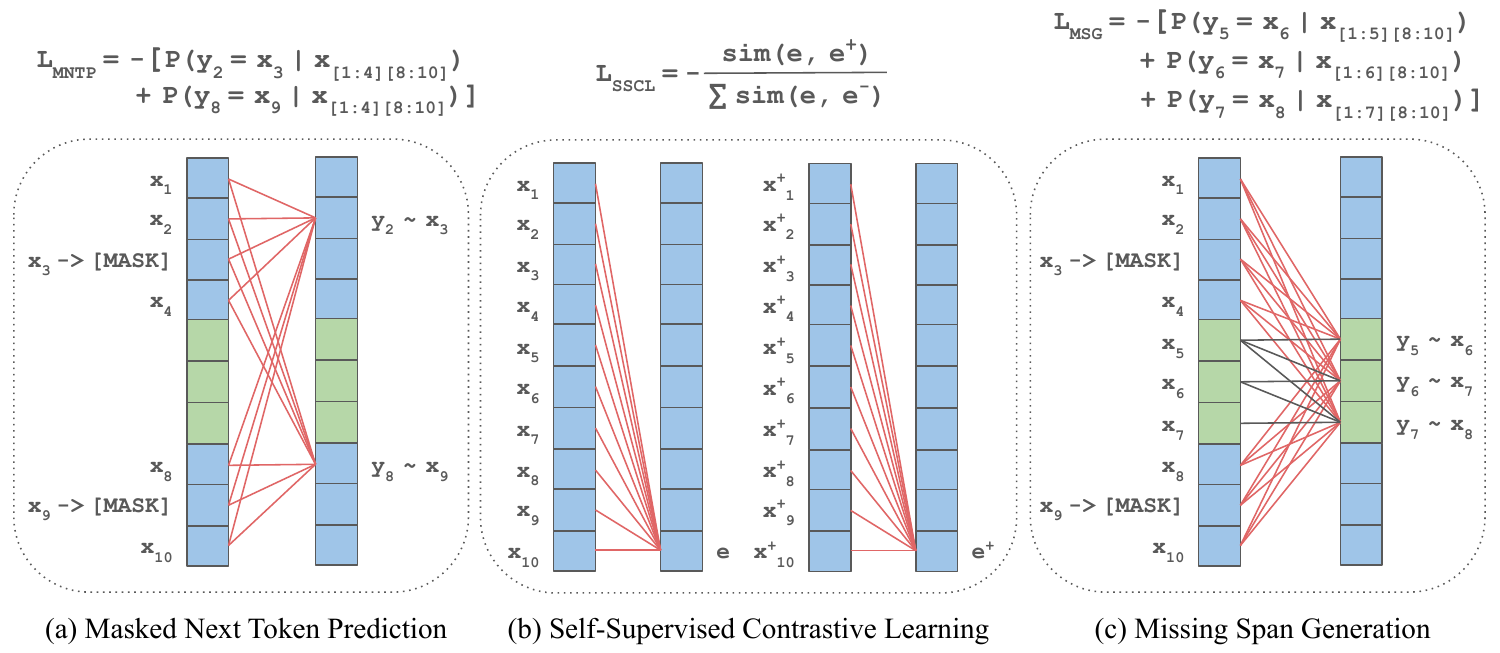}
    \vskip -0.1in
    \caption{\unigr{} training objectives include: (a) Masked next token prediction, which is applied on the output corresponding to the token preceding the masked context token. (b) Self-supervised contrastive learning, which is applied on the model's representation corresponding to the last token. (c) Missing span generation, which is applied on the output corresponding to the span tokens. In this illustration, the red lines denote bidirectional attention and the black lines denote causal attention. Further, for (a) and (c), the output token $y_i$ is trained to predict the input token $x_{i+1}$, as denoted by "$y_i \sim x_{i+1}$"}
    \label{fig:training-objectives}
\end{figure*}

\subsection{Modifying Attention}
\label{sec:modified-attention}
We categorize the input tokens as either \textit{context tokens} or \textit{span tokens} and use the attention mask shown in Figure~\ref{fig:modified-attention}.

\boldheader{Context tokens} Each context token (shown in blue in Figure~\ref{fig:modified-attention}) attends to all other context tokens in the sequence. The attention mask has 0s at output positions corresponding to context tokens, allowing each context token to access information at every other context token. This transformation shifts the original unidirectional LLM into a bidirectional model.

\boldheader{Span tokens} The span tokens (shown in green in Figure~\ref{fig:modified-attention}) are a contiguous span of input tokens that attend to all context tokens and have causal attention among themselves. By enabling span tokens to access surrounding context, we effectively convert the original LLM into an infilling language model. Additionally, the causal attention among span tokens preserves the LLM's generative capabilities, which could be compromised if bidirectionality is fully unlocked (see Section~\ref{sec:open-generation} for details).

During training, an input sequence includes one or more spans of span tokens surrounded by context tokens. During inference, the attention mechanism can operate in three modes: (1) fully causal/unidirectional for open-ended text generation tasks, (2) fully bidirectional representation learning tasks, or (3) a combination of causal and bidirectional for text infilling.

\begin{figure*}[t]
    \centering
    \includegraphics[width=0.95\textwidth]{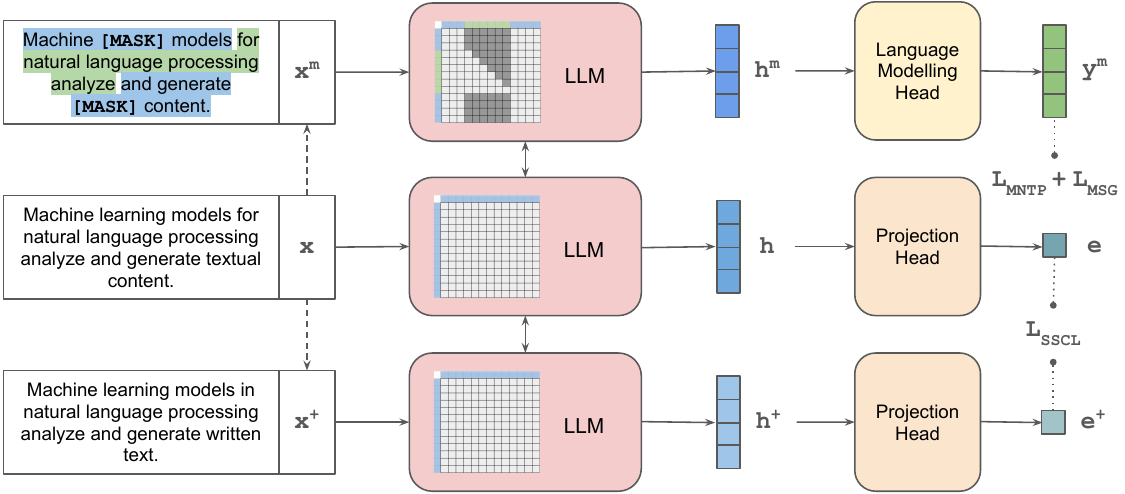}
    \vskip -0.1in
    \caption{\unigr{} processes three views of the input using different attention mechanisms within the same LLM. The model is trained (or fine-tuned) using three self-supervised learning objectives simultaneously to augment it with the ability to generate token-level and sentence-level representations and perform text infilling tasks, while maintaining its original left-to-right text generation capability.}
    \label{fig:overview}
\end{figure*}

\subsection{Training Objectives}
\label{sec:training-objectives}
\unigr{} fine-tunes an off-the-shelf LLM using three self-supervised objectives. 
These objectives are illustrated in Figure~\ref{fig:training-objectives} and discussed below.

\subsubsection{Masked Next Token Prediction (MNTP)} 
MNTP enables the model to realize its newly enabled bidirectional attention capability. The task is defined as follows: Given an input sequence $\x{} = (x_1, x_2, ...,  x_L)$, we select a fraction of the input tokens for masking and train the model to predict these masked tokens. In our setup, we find that selecting 20\% of the input tokens for masking works well. Further, following \cite{Devlin2019BERTPO}, we replace 80\% of the selected tokens with a $\texttt{[MASK]}$ token, 10\% with a random token from the model's vocabulary, and leave the remaining 10\% unchanged. Since LLMs are trained to predict the next token in a sequence, we use the token representations from position $l$ to predict a masked token at position $l+1$ (as shown in Figure~\ref{fig:training-objectives}a). In Appendix~\ref{sec:mtp}, we also explore the possibility of using the standard masked token prediction (MTP) objective, where the output at token $l$ predicts the masked token at position $l$ and find that MTP performs poorly for LLMs that are trained to predict autoregressively. MNTP is optimized using categorical cross-entropy loss:
\begin{align*}
    \mathcal{L}_{\text{MNTP}} = & \frac{-1}{NL} \sum_{n=1}^{N} \sum_{l=1}^{L} \sum_{v=1}^{V} \\
    & \left( \mathbbm{1}_{\text{mask}}(l+1) \cdot (y^{(n)}_{lv} \log(\hat{y}^{(n)}_{lv})) \right)
\end{align*}
where $N$ denotes batch size, $L$ denotes sequence length, $V$ denotes vocabulary size, $\mathbbm{1}_{\text{mask}}(l+1)$ is 1 if position $l+1$ is masked and 0 otherwise, and $y_{lv}$ and $\hat{y}_{lv}$ represent true and predicted probabilities for $v^{th}$ token in vocabulary at position $l$ in the sequence. Note that this task is conducted exclusively with the context tokens.

\subsubsection{Self-Supervised Contrastive Learning (SSCL)}
Since LLMs are not explicitly trained to capture the entire input context and generate sentence-level representations, we employ SSCL to transform them into text encoders. The task is defined as follows: Given an input sequence $\x{}$, we generate its augmented view $\x{}^{\mathbf{+}}$ and align their encoded representations $\mathbf{e} = f(\x{})$ and $\mathbf{e^{\mathbf{+}}} = f(\x{}^{\mathbf{+}})$ in the embedding space, while distancing them from the encodings $\mathbf{e^{\mathbf{-}}} = f(\x{}^{\mathbf{-}})$ of other input sequences $\x{}^{\mathbf{-}}$ in a training batch. Specifically, we employ paraphrasing \citep{prithivida2021parrot} to generate augmented views of an input, and add an instruction "\textit{Given the sentence, find its representation:}" to the training examples \citep{Jiang2023ScalingSE}. Then, we use the output corresponding to the last token ($\texttt{[EOS]}$) of the final hidden states as the sentence encoding. Our choice of using the last token representation as the encoding is guided by the fact that \unigr{} optimizes simultaneously for token-level and sentence-level representations. Since the last token's representation is not used for token-level optimization (because the representation of input token $l$ is given by output token $l-1$), this choice enables us to disentangle the two representation learning tasks during joint training. We use InfoNCE \citep{Oord2018RepresentationLW} with in-batch negatives as the loss function:
$$
    \mathcal{L}_{\text{SSCL}} = \frac{-1}{N} \sum_{i=1}^{N} \log \frac{\exp(\textbf{e}_i \cdot \textbf{e}^+_i / \tau)}{\sum_{j=1}^{N} \exp(\textbf{e}_i \cdot \textbf{e}^-_j / \tau)}
    \label{eq:infonce}
$$
where $N$ represents the batch size and $\tau$ denotes the temperature for logit scaling.

\subsubsection{Missing Span Generation (MSG)}
MSG provides text infilling capabilities to the left-to-right autoregressive model. The task is defined as: Given a position $p$ and an input sequence $\x{} = (x_1, ..., x_p, x_q, ...,  x_L)$, generate a plausible sequence of $m$ tokens $\y{} = (y_1, y_2, ...,  y_m)$ that fits between $x_p$ and $x_q$. More specifically, in our training setup, this task entails predicting a span token $y_l$ conditioned on all context tokens in $\x{}$ and the preceding span tokens $y_{[1..l-1]}$. We train using categorical cross-entropy loss computed over the predicted span tokens:
$$
    \mathcal{L}_{\text{MSG}} = \frac{-1}{N} \sum_{n=1}^{N} \sum_{l=1}^{L} \sum_{v=1}^{V} \mathbbm{1}_{\text{span}}(l) \cdot (y^{(n)}_{lv} \log(\hat{y}^{(n)}_{lv}))
$$
where $N$ denotes batch size, $L$ denotes sequence length, $V$ denotes vocabulary size, $\mathbbm{1}_{\text{span}}(l)$ is 1 if the token at position $l$ is a span token and 0 otherwise, and $y_{lv}$ and $\hat{y}_{lv}$ are the true and predicted probabilities for token $v$ in the vocabulary at position $l$ in the sequence. The standard next token prediction task of LLMs can be considered as a special case of this objective, wherein all input tokens are span tokens (and the attention mechanism reduces to causal attention). Thus, a beneficial side effect of this task is that the model retains its text generation capability while learning bidirectional representations.

\subsection{Approach Overview}
Figure~\ref{fig:overview} provides an overview of \unigr{}. Starting with a training example $\x{}$, the process unfolds in two parallel streams -- (1) One or more contiguous spans of $M$ tokens in $\x{}$ are marked as span tokens, while a fraction of the remaining tokens (context tokens) is masked to form $\x{}^\text{m}$. (2) $\x{}$ is augmented to get $\x{}^{+}$. The input sequences $\x{}$, $\x{}^\text{m}$ and $\x{}^{+}$ are processed by the base decoder model to produce hidden states $\h{}$, $\h{}^\text{m}$ and $\h{}^{+}$. From $\h{}^\text{m}$, a language modeling head generates $\y{}^\text{m}$, which is used to compute $\mathcal{L}_{\text{MNTP}}$ and $\mathcal{L}_{\text{MSG}}$. Parallelly, $\h{}$ and $\h{}^{+}$ are processed using a projection head to get $\textbf{e}$ and $\textbf{e}^{+}$, which are used to compute $\mathcal{L}_{\text{SSCL}}$. The overall loss function is:
$$
    \mathcal{L} = \lambda_1 \mathcal{L}_\text{MNTP} + \lambda_2 \mathcal{L}_\text{SSCL} + \lambda_3 \mathcal{L}_\text{MSG}
    \label{eq:overall-loss}
$$
For processing $\x{}$ and $\x{}^{+}$, the decoder uses a bidirectional attention mask (as shown in Figure~\ref{fig:bidirectional-mask}). For processing $\x{}^\text{m}$, the decoder employs an attention mask similar to those depicted in Figures~\ref{fig:unigr-mask-1} and \ref{fig:unigr-mask-2}. In some cases, when all input tokens are marked as span tokens, the attention mask reduces to causal attention, as shown in Figure~\ref{fig:causal-mask}.
\section{Experiments}
\label{sec:experiments}
In this section, we demonstrate that \unigr{} enhances a decoder-only LLM with representation learning and infilling capabilities while preserving its original generative abilities. Specifically, we show that LLMs adapted with \unigr{} outperform the base model and other adaptation methods on representation learning tasks (Sections~\ref{sec:word-task} and \ref{sec:sentence-task}). We then highlight how \unigr{} significantly improves the LLM’s ability to infill missing spans (Section~\ref{sec:infilling-task}). Finally, we analyze the effect of \unigr{} on the original generative and reasoning capabilities of the LLM (Section~\ref{sec:knowledge-retention}). All training details are mentioned in Appendix~\ref{sec:training-details}. Additionally, we present ablation experiments demonstrating the benefits of training a bidirectional language model with a causal objective in Appendix~\ref{sec:objective-ablation}.

It is to be noted that our goal is not to achieve state-of-the-art results on a specific benchmark. Instead, we aim to enhance a pretrained LLM with additional capabilities while preserving its original performance. Therefore, our main baselines are the base LLM and other methods that augment the same LLM with specific capabilities.

\begin{table}[tp]
    \centering
    \small
    \begin{tabular}{c c c c}
        \toprule
        \textbf{Model} & \textbf{Chunking} & \textbf{NER} & \textbf{POS-Tags} \\
        \midrule
        \multicolumn{4}{c}{\textit{Encoder models}} \\
        \midrule
        BERT-Large & 71.77 & 90.09 & 75.12 \\
        XLNet-Large & 79.70 & 93.67 & 83.02 \\
        DeBERTa-Large & 85.74 & 94.97 & 86.49 \\
        StructBERT-Large & 89.99 & 97.31 & 90.86 \\
        \midrule
        \multicolumn{4}{c}{\textit{Llama 2 models}} \\
        \midrule
        Llama-2-7B & 88.23 & 96.59 & 91.53 \\
        LLM2Vec & 89.66 & 96.05 & 90.53 \\
        $\text{LLM2Vec}^\text{[MNTP]}$ & 91.61 & 97.16 & 92.61 \\
        \unigr{} & \textbf{92.64} & \textbf{98.31} & \textbf{93.34} \\
        \bottomrule
    \end{tabular}
    \vskip -0.05in
    \caption{Results on word-level tasks. LLM2Vec \citep{BehnamGhader2024LLM2VecLL} adapts the model using MNTP and SimCSE. $\text{LLM2Vec}^\text{[MNTP]}$ is an intermediate state of LLM2Vec that is trained only on MNTP. All numbers except those for \unigr{} are taken from \cite{BehnamGhader2024LLM2VecLL}.}
    \vskip -0.15in
    \label{tab:word-tasks}
\end{table}
\begin{table*}[t]
    \centering
    \small
    \begin{tabular}{c c c c c c c c c}
        \toprule
        \textbf{Model} & \textbf{STS12} & \textbf{STS13} & \textbf{STS14} & \textbf{STS15} & \textbf{STS16} & \textbf{STS-B} & \textbf{SICK-R} & \textbf{Avg} \\
        \midrule
        \multicolumn{9}{c}{\textit{Encoder models (finetuned using SimCSE)}} \\
        \midrule
        BERT-Base & 68.40 & 82.41 & 74.38 & 80.91 & 78.56 & 76.85 & 72.23 & 76.25 \\
        RoBERTa-Base & 70.16 & 81.77 & 73.24 & 81.36 & 80.65 & 80.22 & 68.56 & 76.57 \\
        RoBERTa-Large & \textbf{72.86} & 83.99 & 75.62 & \textbf{84.77} & \textbf{81.80} & 81.98 & 71.26 & 78.90 \\
        \midrule
        \multicolumn{9}{c}{\textit{Llama 2 models}} \\
        \midrule
        Llama-2-7B & 50.98 & 74.02 & 62.86 & 67.09 & 71.03 & 63.56 & 67.22 & 65.25 \\
        Echo Embeddings & 52.40 & 72.40 & 61.24 & 72.67 & 73.51 & 65.73 & 64.39 & 66.05 \\
        LLM2Vec & 65.39 & 79.26 & 72.98 & 82.72 & 81.02 & 78.32 & 71.77 & 75.92  \\
        \unigr{} & 67.98 & \textbf{84.66} & \textbf{77.67} & 84.17 & 79.44 & \textbf{82.88} & \textbf{78.77} & \textbf{79.36} \\
        \bottomrule
    \end{tabular}
    \vskip -0.05in
    \caption{Results on STS tasks. The encoder models are trained using SimCSE and their results are taken from \citet{Gao2021SimCSESC}. The results for Llama-2-7B are obtained using the last token embedding from the final hidden state as the sentence representation. The results for LLM2Vec and Echo Embeddings are taken from \citet{BehnamGhader2024LLM2VecLL} and \citet{Springer2024RepetitionIL}, respectively.}
    \vskip -0.05in
    \label{tab:sts-tasks}
\end{table*}
\begin{table*}[t]
    \centering
    \small
    \begin{tabular}{c c c c}
        \toprule
        \textbf{Dataset} & \textbf{BiorxivClustering} & \textbf{TwentyNewsgroups} & \textbf{MedrxivClustering} \\
        \midrule
        Echo Embeddings & 25.92 & 23.42 & 24.30 \\ 
        LLM2Vec & 34.69 & 30.76 & 29.49 \\ 
        \unigr{} & \textbf{35.10} & \textbf{53.31} & \textbf{30.21} \\ 
        \bottomrule
    \end{tabular}
    \vskip -0.05in
    \caption{Results on clustering tasks. The results for LLM2Vec and Echo Embeddings are taken from \cite{BehnamGhader2024LLM2VecLL} and \cite{Springer2024RepetitionIL}, respectively.}
    \vskip -0.15in
    \label{tab:clustering-tasks}
\end{table*}

\subsection{Word-Level Tasks}
\label{sec:word-task}
We evaluate the token-level representations on three tasks -- (1) chunking, (2) named entity recognition, and (3) part-of-speech tagging -- using the CoNLL-2003 dataset \citep{TjongKimSang2003IntroductionTT}. After applying the training objectives proposed in Section~\ref{sec:training-objectives}, we train a linear classifier on top of the frozen representations obtained from the last hidden state of the model. The word-level embeddings are obtained by averaging the representations of the tokens that make up that word. Further, the representation of the token at position $i$ is given by the embedding at position $i-1$. 

Table~\ref{tab:word-tasks} compares \unigr{} with powerful encoder models and LLM2Vec \citep{BehnamGhader2024LLM2VecLL}, a recent method for adapting decoder-only LLMs for representation learning. The second-best approach, $\text{LLM2Vec}^\text{[MNTP]}$, relies solely on MNTP for model adaptation. In contrast, \unigr{} integrates both representation learning objectives (MNTP and SSCL) and generative objectives (MSG). The superior performance of \unigr{} over $\text{LLM2Vec}^\text{[MNTP]}$, despite using the same training data, model, and parameters, highlights the synergistic advantages of a unified training strategy for word-level representation learning.


\subsection{Sentence-Level Tasks}
\label{sec:sentence-task}
We evaluate sentence-level representations on multiple semantic similarity and clustering benchmarks \citep{Muennighoff2022MTEBMT}. We perform these tasks using the representation corresponding to the last token (\texttt{[EOS]}), without performing any task-specific training. Further, task-specific instructions (Table~\ref{tab:task-instructions}) are used for extracting relevant representations \citep{Su2022OneEA, Wang2023ImprovingTE}.

We compare the text encoding capabilities of \unigr{} with other recently proposed methods for transforming decoder models into text encoders, viz. LLM2Vec \citep{BehnamGhader2024LLM2VecLL} and Echo Embeddings \citep{Springer2024RepetitionIL}. Table~\ref{tab:sts-tasks} shows the results on Semantic Textual Similarity (STS) task and Table~\ref{tab:clustering-tasks} shows the results on clustering tasks. As can be seen, \unigr{} outperforms other adaptation methods on STS and clustering tasks. As previously noted, the fact that \unigr{} surpasses LLM2Vec suggests the potential benefit of a unified training approach. 

\begin{table}[t]
    \centering
    \small
    \begin{tabular}{c c c}
        \toprule
        \textbf{Method} & \textbf{ROC Stories} & \textbf{Wikitext-103} \\
        \midrule
        Llama-2-7B & 13.9347 & 22.0399 \\
        \unigr{} & \textbf{9.5161} & \textbf{15.4573} \\
        \bottomrule
    \end{tabular}
    \vskip -0.05in
    \caption{Results on the infilling tasks. We measure the perplexity (PPL) for sentence infilling and block-of-text infilling on ROC-Stories and Wikitext-103, respectively.}
    \vskip -0.15in
    \label{tab:infilling-ppl}
\end{table}

\begin{table}[t]
    \centering
    \small
    \begin{tabular}{c c}
        \toprule
        \textbf{Method} & \textbf{Score} \\
        \midrule
        Unidirectional Llama-2-7B & 53.5 \\
        Zero-Shot Setup & 5.5 \\
        Five-Shot Setup & 54.5 \\
        \unigr{} & \textbf{62.0} \\
        \bottomrule
    \end{tabular}
    \vskip -0.05in
    \caption{Human evaluation for the infilling tasks. The score denotes the percentage of infillings that were considered contextually appropriate by human evaluators.}
    \vskip -0.15in
    \label{tab:infilling-human-eval}
\end{table}

\subsection{Infilling Task}
\label{sec:infilling-task}
To test infilling capabilities, we evaluate the perplexity (PPL) of Llama-2-7B and \unigr{}-adapted Llama-2-7B on the test set of ROC Stories \citep{Mostafazadeh2016ACA} and Wikitext-103 \citep{Merity2016PointerSM}. For ROC Stories, we randomly mask out a sentence from each 5-sentence story, while for Wikitext-103, we mask up to three spans with lengths ranging from 8 to 32 tokens. Following \cite{Donahue2020EnablingLM}, we compute PPL only for the tokens comprising the original masked out spans. The results are presented in Table~\ref{tab:infilling-ppl}, and they show that the base model (Llama-2-7B) exhibits significantly higher perplexity for the masked spans compared to \unigr{}, demonstrating that \unigr{} effectively augments the base model with text infilling capabilities. 

We also conduct experiments using zero-shot and few-shot learning to enable Llama-2-7B to incorporate all the surrounding context when infilling a missing span. We explore various prompting strategies and found that while a zero-shot setup did not yield sensible infillings, a five-shot setup with descriptive prompts resulted in more context-aware infillings (refer to Appendix~\ref{sec:prompt-infilling} for details). For a comprehensive analysis, we conducted a human evaluation to compare the quality of infillings generated by the base model, its zero-shot variant, its few-shot variant, and its \unigr{} adaptation. In this evaluation, we randomly sampled 100 stories from the ROC Stories dataset, masked out one of their middle sentences, and tasked the models with infilling the missing sentence. Two human annotators on Amazon Mechanical Turk (with at least a high school diploma) then independently assessed whether each generated sentence was contextually appropriate and contributed to a coherent story. The results are presented in Table~\ref{tab:infilling-human-eval}, showing that the infillings generated by \unigr{}-adapted model are more coherent than those generated by the base model. We show some qualitative examples of infilling in Table~\ref{tab:infilling-examples}.

\begin{table}[t]
    \centering
    \small
    \captionsetup{type=table}
    \begin{tabular}{c c c c c}
        \toprule
        \multirow{2}{*}{\textbf{Method}} & \multicolumn{2}{c}{\textbf{Wikitext-103}} & \multicolumn{2}{c}{\textbf{ROC Stories}} \\
         & \textbf{Rep-Sen} & \textbf{Rep-4} & \textbf{Rep-Sen} & \textbf{Rep-4} \\
        \midrule
        Llama-2-7B & 0.0056 & 0.0601 & 0.0381 & 0.0163 \\
        LLM2Vec & 0.2044 & 0.4747 & 0.2945 & 0.5243 \\
        \unigr{} & 0.0151 & 0.2047 & 0.0737 & 0.2573 \\
        \bottomrule
    \end{tabular}
    \vskip -0.05in
    \caption{Analyzing the repetition problem. Both LLM2Vec and \unigr{} are applied for 3400 iterations.}
    \vskip -0.15in
    \label{tab:repetition-problem}
\end{table}

\begin{figure}[b]
    \centering
    \includegraphics[width=0.45\textwidth]{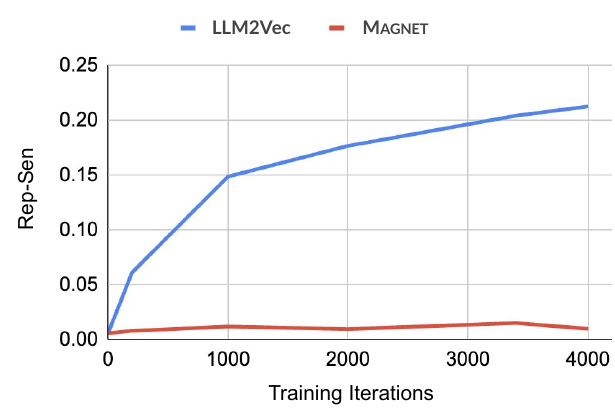}
    \vskip -0.15in
    \caption{LLM2Vec increases text repetition with more training, while no such trend is observed for \unigr{}.}
    \label{fig:repetition-problem}
\end{figure}
\begin{table*}[t]
    \centering
    \small
    \begin{tabular}{c c c c c c c c c c}
        \toprule
        \multirow{2}{*}{\textbf{Model}} & \multirow{2}{*}{\textbf{HellaSwag}} & \multirow{2}{*}{\textbf{BBH}} & \multicolumn{2}{c}{\textbf{ARC}} & \multirow{2}{*}{\textbf{NQ}} & \multicolumn{4}{c}{\textbf{MMLU}} \\
        \cmidrule(lr){4-5}
        \cmidrule(lr){7-10}
        & & & \textbf{Easy} & \textbf{Challenge} & & \textbf{Humanities} & \textbf{STEM} & \textbf{Social Science} & \textbf{Other} \\
        \midrule
        Llama-2-7B & 75.51 & 33.57 & 73.95 & 44.28 & 24.02 & 43.27 & 36.09 & 53.04 & 54.84 \\
        \unigr{} & 75.08 & 32.22 & 74.33 & 44.52 & 24.22 & 42.25 & 36.63 & 52.64 & 52.40 \\
        \bottomrule
    \end{tabular}
    \vskip -0.05in
    \caption{Evaluating the impact of \unigr{} on Llama-2-7B's performance across benchmarks. The metrics are computed using the LM Evaluation Harness~\citep{evalharness}. Due to the undisclosed evaluation prompts for Llama, reproducing the exact baseline results is difficult. We adopt the same setup for both Llama-2-7B and \unigr{}.}
    \vskip -0.15in
    \label{tab:mmlu}
\end{table*}

\subsection{Repetition Problem}
\label{sec:open-generation}
The repetition problem in text generation refers to the issue when generative models repeatedly produce the same phrases or sentences. Prior studies have identified that this issue often results from biases in the training data, limitations in the model's design, or standard likelihood training and inference \citep{Holtzman2019TheCC, Welleck2019NeuralTG, Fu2020ATA, Xu2022LearningTB}. In our study, we find that when generative decoder models are adapted into text encoders by enabling bidirectional attention \citep{BehnamGhader2024LLM2VecLL, Li2023BeLLMBD, Li2023LabelSL}, the issue of repetition is significantly worsened. For example, Table~\ref{tab:generation-examples} shows the texts generated (using greedy decoding) by the original Llama-2-7B and its LLM2Vec adaptation \citep{BehnamGhader2024LLM2VecLL}. We observe noticeable repetitions in the text generated by LLM2Vec-adapted-LLaMA, although the fine-tuning data (Wikitext-103) had almost no sentence-level repetitions (0.02\%). 

For quantitative detection of text repetitions, we compute $\textit{Rep-Sen} = 1.0 - \frac{\mid\text{unique sentences}\mid}{\mid\text{sentences}\mid}$ and $\textit{Rep-n} = 1.0 - \frac{\mid\text{unique n-grams}\mid}{\mid\text{n-grams}\mid}$, as done by prior works analyzing the repetition problem \citep{Holtzman2019TheCC, Welleck2019NeuralTG, Xu2022LearningTB}. Specifically, we create a \textit{prefix-dataset} from the test sets of Wikitext-103 and ROC Stories, consisting of 5-word and single-sentence prefixes, respectively. The model is then tasked with autoregressively generating text based on these prefixes. Table~\ref{tab:repetition-problem} shows the repetition metrics for Llama-2-7B and its adaptations using LLM2Vec and \unigr{}. As can be seen, in comparison to LLM2Vec, \unigr{} makes the base model significantly less prone to repeating sentences. For instance, for Wikitext-103, LLM2Vec makes Llama-2-7B 36.5 times more likely to repeat sentences, while \unigr{} only makes it 2.7 times more likely. Further, as shown in Figure~\ref{fig:repetition-problem}, the repetition problem exacerbates with additional iterations of LLM2Vec training, whereas no similar trend is observed with \unigr{}.

We conjecture that LLM2Vec is significantly more prone to generating repetitive text because it exclusively focuses on learning representations with bidirectional attention. This training approach perhaps makes the decoder model somewhat similar to bidirectional LMs like BERT, which are known to repeat words when used for text generation (Table~\ref{tab:generation-examples}). \unigr{} solves this issue by having autoregressive generation as an objective.

\subsection{Knowledge and Reasoning Tasks}
\label{sec:knowledge-retention}
We assess the effect of \unigr{} on the knowledge and reasoning capabilities acquired by the LLM during pretraining. Specifically, we evaluate its performance on HellaSwag (0-shot)~\citep{Zellers2019HellaSwagCA}, BBH (3-shot)~\citep{Suzgun2022ChallengingBT}, ARC (0-shot)~\citep{Clark2018ThinkYH}, MMLU (5-shot)~\citep{hendryckstest2021}, and NaturalQuestions (5-shot)~\cite{Kwiatkowski2019NaturalQA}, covering commonsense reasoning and world knowledge.

For this evaluation, we fine-tune the model on the SlimPajama dataset~\citep{cerebras2023slimpajama}, which includes diverse text sources such as CommonCrawl, C4, GitHub, Books, ArXiv, Wikipedia, and StackExchange. This choice ensures that the fine-tuning data resembles Llama-2-7B’s original pretraining distribution (despite its exact composition being unknown). By doing so, we mitigate potential biases introduced by highly structured datasets like Wikitext, which could favor Wikipedia-derived tasks such as NaturalQuestions while disadvantaging commonsense reasoning benchmarks like HellaSwag.

As shown in Table~\ref{tab:mmlu}, \unigr{} has minimal impact on the model's knowledge and reasoning capabilities. The minor variations observed can be attributed to differences in dataset composition during fine-tuning. Furthermore, to ensure a comprehensive evaluation, we assess the performance of the model fine-tuned with SlimPajama on representation learning tasks and find the results consistent with the metrics reported in Tables~\ref{tab:word-tasks} and \ref{tab:sts-tasks} (where Wikitext-103 was used for fine-tuning). Specifically, when fine-tuned with SlimPajama, the model attains 92.00\% on chunking, 98.30\% on NER, 93.21\% on POS-tagging, and an average of 79.33\% on the STS tasks.
\section{Conclusion}
\label{sec:conclusion}
In this work, we presented \unigr{}, a method to transform causal LLMs into text encoders and infilling language models with bidirectional context-capturing ability. Through extensive experiments, we show that \unigr{} uniquely equips LLMs with abilities that are beyond the scope of traditional text encoders or decoders. Thus, \unigr{} shows the potential to unify text generation and text encoding within a single framework. 

\bibliography{main}

\clearpage
\appendix
\section{Training Details}
\label{sec:training-details}
\unigr{} fine-tunes Llama-2-7B using LoRA \citep{Hu2021LoRALA} with $r=16$ and $\alpha=32$. We use the AdamW optimizer with $\beta_1=0.9$, $\beta_2=0.999$ and $\epsilon=1e-8$, apply bfloat16 quantization, and use scaled-dot-product attention (SDPA). All experiments are performed on a single NVIDIA A100 GPU, with the \unigr{} adaptation of Llama-2-7B taking approximately 7 hours. We discuss the training dataset and hyperparameters for the different objectives/tasks below.

\boldheader{Datasets} For representation learning and infilling tasks, we train on Wikitext-103~\citep{Merity2016PointerSM} to ensure a fair comparison with our baselines in Sections~\ref{sec:word-task} and \ref{sec:sentence-task}. For knowledge and reasoning tasks (Section~\ref{sec:knowledge-retention}), we use SlimPajama~\citep{cerebras2023slimpajama} to mitigate biases from highly structured datasets like Wikitext, which could favor Wikipedia-derived tasks. The WikiText-103 dataset is released under the Creative Commons Attribution-ShareAlike license, and SlimPajama is available under the Apache 2.0 license, both permitting use in open-source research.

\boldheader{MNTP} We train for 4200 iterations using a batch size of 32, a learning rate of 3e-5, and a max sequence length of 512. We select 20\% of the tokens for masking -- 80\% of the selected tokens are replaced with a \texttt{[MASK]} token, 10\% tokens are replaced with a random token from the model's vocabulary, and 10\% tokens are left unchanged. For Llama-2-7B, we use "\textunderscore" as the mask token. 

\boldheader{SSCL} We train for 800 iterations using a batch size of 64, a learning rate of 3e-5, and a max sequence length of 128. To extract representations we use the prompt "\textit{Given the sentence, find its representation:}" and extract the representations corresponding to the last token. The training data is created by extracting lines longer than 20 words and paraphrasing them for the positive examples. We set $\tau=0.1$ in equation \ref{eq:infonce}.

\boldheader{MSG} Similar to MNTP, we train for 4200 iterations using a batch size of 32, a learning rate of 3e-5, and a max sequence length of 512. A training example can have up to 2 missing spans, with span length ranging from 4 to 128 tokens.

\boldheader{Overall Loss} For the first 3400 iterations, we optimize the loss (equation \ref{eq:overall-loss}) with $\lambda_1 = 1$, $\lambda_2 = 0$, and $\lambda_3 = 1$, and for the next 800 iterations $\lambda_1 = 1$, $\lambda_2 = 9$, and $\lambda_3 = 1$. Initially, we train with only MNTP and MSG, as these objectives help the model learn to capture future context---a capability the base model lacks. However, this choice mainly contributes to faster training, as similar results are obtained when training with all objectives from the start.

\boldheader{Word-Level Tasks} Using the frozen representations from the last hidden layer of the base model, we train a linear classifier for the three word-level tasks (Chunking, NER, and POS-tagging). Specifically, we train on the CoNLL-2003 train set for 4000 steps using a batch size of 8, a learning rate of 5e-4, and a dropout rate of 0.1.

\section{Contextual Prompt Infilling}
\label{sec:prompt-infilling}
To thoroughly evaluate the infilling capability of the base model, we perform zero-shot and few-shot experiments where the model is shown both preceding and following context of a missing span of text. 

\subsection{Zero-Shot Evaluation}
To this end, we experimented with four types of prompts to infill a missing line in five-line stories from the ROC Stories dataset. The four prompting strategies we used are:

\boldheader{Blank Infilling Prompt} In this setting, we add a blank token (\textunderscore) at the infilling position and use the following prompt: 
\begin{center}
    \textit{Generate the missing line represented by \textunderscore\; in the given text: \textless text\textgreater. \\
    Generate a single sentence.\\
    The missing line is: }
\end{center}
Here, \textit{\textless text\textgreater} represents the input text with "\textunderscore" in place of a missing sentence.

\boldheader{Contextual Prompt} In this setting, we provide the past and future context of the missing line and use the following prompt: 
\begin{center}
    \textit{Fill in the missing sentence between "\textless past-context\textgreater" and "\textless future-context\textgreater". Generate only one sentence. The missing sentence is: }
\end{center}

\boldheader{Prefix-Suffix Prompt} In this setting, we give the past context of a missing sentence as a prefix and the future context as a suffix and ask the model to generate the middle. Specifically, we use the following prompt: 
\begin{center}
    \textit{Given the prefix and the suffix, generate the middle sentence. \\
    Prefix: \textless past-context\textgreater. \\ Suffix: \textless future-context\textgreater. \\ Generate only one sentence. \\ Middle: }.
\end{center}

\boldheader{Line-by-Line Prompt} In this setting, we make the prompt more descriptive by providing all the available context, specifying the line number for all the available lines, and asking for the missing line. For instance, if the task is to infill the second line of a five-sentence story, the prompt would be:
\begin{center}
    \textit{You have a five-sentence story with some missing text. \\
    Here is the context for each line, with the missing line indicated:  \\
    Line 1: \textless line-1\textgreater \\ Line 2: [Missing Line] \\ Line 3: \textless line-3\textgreater \\ Line 4: \textless line-4\textgreater \\ Line 5: \textless line-5\textgreater \\
    Please generate the missing line of the story. Please generate only the missing line and nothing else.\\ The missing line is: Line 2: }
\end{center}

For the abovementioned prompting strategies, we experimented with various prompt variations, including paraphrasing the instructions, using "\texttt{[MASK]}", "\texttt{[blank]}" or "\textunderscore" to denote the missing line, and addressing common avoidable errors using the instructions (for e.g., adding "\textit{Generate only one line.}" to enforce single line infillings and avoid formatting issues). In general, we find that regardless of the prompting strategy used, Llama-2-7B repeats/paraphrases one of the provided lines or summarizes the context as the infilling. In some cases it even ends up generating totally random text (like code). This is perhaps because the model is not trained for the infilling task. Table~\ref{tab:zero-shot-prompt-infilling} shows some qualitative examples of text infilling using the different prompting methods.

\subsection{Few-Shot Evaluation}
To improve infilling results from the base model, we employed few-shot learning techniques with various prompting styles -- blank infilling, prefix-suffix, and line-by-line. Specifically, we provided five solved examples in the model's context using the chosen prompt format and asked the model to infill the missing line in the sixth example. We observed that more descriptive prompts and examples led to better output from the model, and the line-by-line prompting style seemed to be the most effective in enabling coherent infillings. We present qualitative examples of the infilling generated using this approach in Table~\ref{tab:infilling-examples}.

\begin{table*}[t]
    \centering
    \small
    \begin{tabular}{c c c c c c c c c}
        \toprule
        \textbf{Training Objectives} & \textbf{Chunking} & \textbf{NER} & \textbf{POS} & \textbf{STS12} & \textbf{STS13} & \textbf{STS14} & \textbf{STS15} & \textbf{STS16} \\
        \midrule
        MNTP & 92.44 & 98.11 & 93.18 & -- & -- & -- & -- & -- \\
        SSCL & -- & -- & -- & \textbf{69.06} & 84.53 & \textbf{78.07} & 84.09 & 78.52 \\
        MNTP + MSG & 92.51 & 98.20 & \textbf{93.38} & -- & -- & -- & -- & -- \\
        SSCL + MSG & -- & -- & -- & 68.46 & 84.52 & 77.33 & \textbf{84.35} & 79.17 \\
        MNTP + SSCL + MSG & \textbf{92.64} & \textbf{98.31} & 93.34 & 67.98 & \textbf{84.66} & 77.67 & 84.17 & \textbf{79.44} \\
        \bottomrule
    \end{tabular}
    \vskip -0.05in
    \caption{Ablation analysis of the proposed training objectives to assess the potential benefits of unified training.}
    \vskip -0.15in
    \label{tab:objective-ablation}
\end{table*}

\section{Training Objective Ablation Analysis}
\label{sec:objective-ablation}
We perform ablation experiments to evaluate the effectiveness of our unified training with the three proposed objectives. Specifically, we compare the performance on representation learning tasks after adapting the LLM using different combinations of the objectives. The results are presented in Table~\ref{tab:objective-ablation}. We find that while MNTP is the only objective that explicitly trains the model for better token-level representations, adding MSG marginally improves performance on word-level tasks. We conjecture that MSG, being closer to the original next-token prediction objective of the base LLM, acts as a regularizer and helps prevent extreme variations in the token representations produced by the model. For sentence-level tasks, which use the SSCL objective on the last token's representation, we observe no noticeable benefit or drawback from including MNTP and MSG. This shows that we can add token-level representation learning and infilling capabilities to the model without hampering performance on sentence-level tasks. We conjecture that the effects of unified training on sentence-level tasks are not evident from Table~\ref{tab:objective-ablation} due to the separation of sentence-level representation learning from token-level representation learning and generation, achieved by using only the last token’s output as the sentence encoding.


\begin{figure*}[t]
    \centering
    \includegraphics[width=0.91\textwidth]{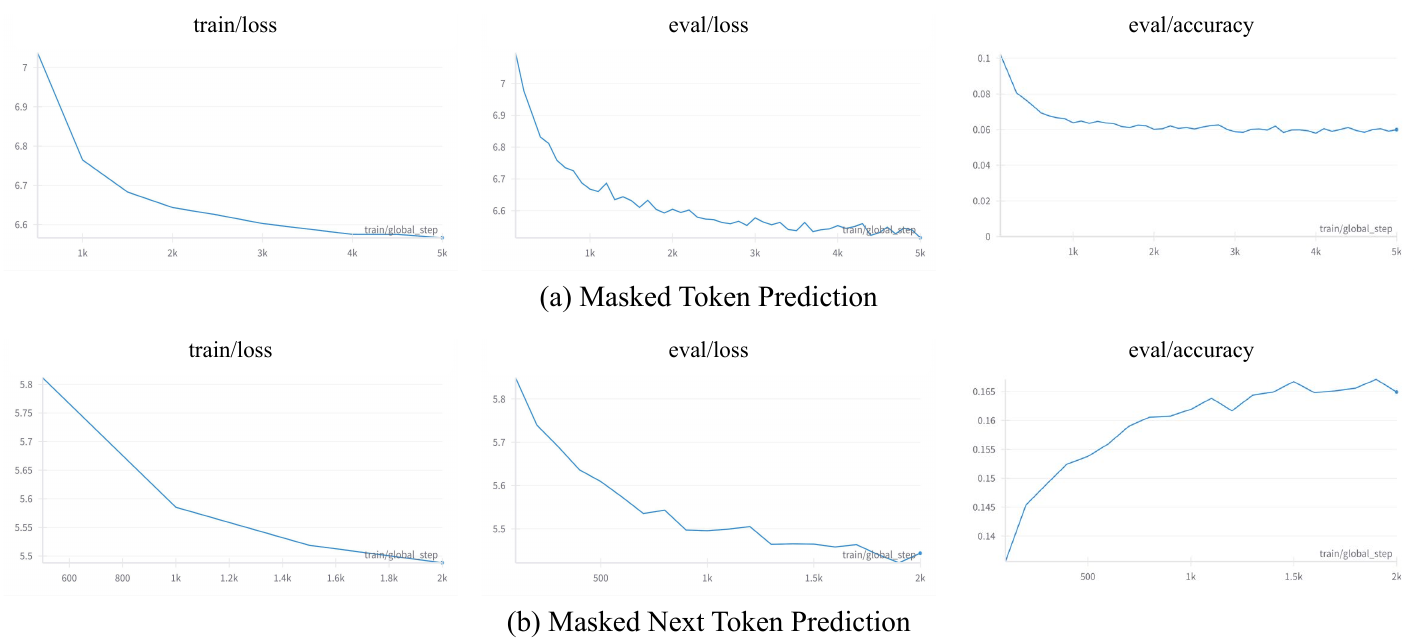}
    \caption{Training curves for MTP and MNTP objectives. When using MTP, model's performance on masked token prediction (measured using eval set accuracy) deteriorates over training iterations.}
    \label{fig:mtp-vs-mntp}
\end{figure*}

\begin{table*}[t]
    \centering
    \small
    \begin{tabular}{c | c}
        \toprule
        \textbf{Task} & \textbf{Instruction} \\
        \midrule
        STS* & Retrieve semantically similar text \\
        BiorxivClusteringP2P & Identify the main category of Biorxiv papers based on the titles and abstracts \\
        BiorxivClusteringS2S & Identify the main category of Biorxiv papers based on the titles \\
        TwentyNewsgroupsClustering & Identify the topic or theme of the given news articles \\
        MedrxivClusteringP2P & Identify the main category of Medrxiv papers based on the titles and abstracts \\
        MedrxivClusteringS2S & Identify the main category of Medrxiv papers based on the titles \\
        \bottomrule
    \end{tabular}
    \vskip -0.05in
    \caption{Instructions used for getting sentence encoding for the different sentence-level tasks. "STS*" refers to all the STS tasks.}
    \vskip -0.15in
    \label{tab:task-instructions}
\end{table*}

\section{Comparing MTP and MNTP Objectives}
\label{sec:mtp}
Traditionally, language models for representation learning are trained to predict a masked token at position $l$ using the output at position $l$ in the final hidden states \citep{Devlin2019BERTPO, Liu2019RoBERTaAR, Lan2019ALBERTAL, Sanh2019DistilBERTAD}. This approach is logical because the residual connections in the transformer block incorporate the $l^{th}$ token's input representation into its output representation. 

We conducted an experiment to test whether we can use LoRA to adapt the base LLM for $l$-to-$l$ prediction (similar to BERT). The training curves for masked token prediction (MTP) and masked number token prediction (MNTP) are shown in Figure~\ref{fig:mtp-vs-mntp}. As illustrated, with MTP, the loss converges, but the evaluation accuracy for masked token prediction decreases. This likely occurs because the base model is trained to predict the $(l+1)^{th}$ token at position $l$, and shifting to $l$-to-$l$ prediction introduces a significant distributional shift that the model may struggle to accommodate swiftly. Thus, overall, we find that MNTP is a more effective objective for converting a decoder-only LLM into a text encoder. Additionally, MNTP aligns well with the causal MSG objective and paves the way for a unified text generator and encoder.

\section{Limitations}
\label{sec:limitations}
Given Llama-2-7B's undisclosed pre-training data composition, there is a potential risk of test set contamination. We mitigate the undue influence of data contamination by benchmarking against the base model and other Llama-2-7B adaptations (LLM2Vec and Echo Embeddings). Future work should establish benchmarks guaranteed to be excluded from LLM pre-training data.

While \unigr{} preserves open-ended generation better than other bidirectional adaptation methods, it does impact the generation quality. For instance, fine-tuning Llama-2-7B with \unigr{} increases Wikitext-103 test-set perplexity from 6.4 to 7.6, indicating slightly reduced next-token prediction confidence despite maintaining artifact-free generation.

For infilling tasks, we focus solely on enabling the base LLM to leverage surrounding context for coherent text completion. The \unigr{}-adapted LLM shows reduced performance when infilling lengthy (more than 128 tokens) mid-text sequences. This limitation can be addressed by: chunking long infills into smaller segments, modifying MSG parameters, or task-specific fine-tuning for infilling.

\begin{table*}[t]
    \centering
    \small
    \begin{tabular}{c | C{4.25in}} 
        \toprule
        \textbf{Method} & \textbf{Infilled Line} \\
        \midrule
        \midrule
        \multicolumn{2}{C{5.25in}}{\textit{John was at the gym. \blank He couldn't get the barbell off his chest. He was too embarrassed to call for help. John suffocated as the weights crushed his chest.}} \\
        \midrule
        \textbf{Blank Infilling} & John was at the gym. \\
        \midrule
        \textbf{Contextual} & John was at the gym, trying to lift a particularly heavy barbell, when he suddenly found himself unable to breathe and his face turned blue. \\
        \midrule
        \textbf{Prefix-Suffix} & John suffocated as the weights crushed his chest. \\
        \midrule
        \textbf{Line-by-Line} & John was at the gym, sweating profusely. \\
        \midrule
        \midrule
        \multicolumn{2}{C{5.25in}}{\textit{Toni took a trip to the local fair with her family. \blank The balloon was red and star shaped. Unfortunately it slipped out of her hands. She was sad to see it go.}} \\
        \midrule
        \textbf{Blank Infilling} & " She watched as it soared high up in the sky."tt = 0 while t < 10:  t += 0. \\
        \midrule
        \textbf{Contextual} & Toni took a trip to the local fair with her family and won a red and star-shaped balloon as a prize. \\
        \midrule
        \textbf{Prefix-Suffix} & Toni took a trip to the local fair with her family and was excited to see a red star-shaped balloon. \\
        \midrule
        \textbf{Line-by-Line} & Toni took a trip to the local fair with her family. \\
        \midrule
        \midrule
        \multicolumn{2}{C{5.25in}}{\textit{I was resting on my couch. \blank They raised their water guns. I immediately ran. I went outside and sprayed them with the hose when they followed.}} \\
        \midrule
        \textbf{Blank Infilling} & They raised their water guns at me. \\
        \midrule
        \textbf{Contextual} & I heard the sound of water splashing and knew they were coming for me. \\
        \midrule
        \textbf{Prefix-Suffix} & I immediately ran. \\
        \midrule
        \textbf{Line-by-Line} & I was resting on my couch, bored out of my mind. \\
        \midrule
        \midrule
        \multicolumn{2}{C{5.25in}}{\textit{Kaylee owned a ferret who always ran free around the house. \blank She put flyers all over the neighborhood. There was a knock at the door three days later. Someone found her ferret.}} \\
        \midrule
        \textbf{Blank Infilling} & She was overjoyed to see her ferret wagging its tail when she opened the door. \\
        \midrule
        \textbf{Contextual} & Kaylee put flyers all over the neighborhood. \\
        \midrule
        \textbf{Prefix-Suffix} & Kaylee found her ferret three days after putting flyers all over the neighborhood. \\
        \midrule
        \textbf{Line-by-Line} & She found the ferret hiding under the bed. \\
        \midrule
        \midrule
        \multicolumn{2}{C{5.25in}}{\textit{Cody really wanted a smart phone. \blank Cody protested that all of his friends had their own phones. His parents phoned some of the other parents. His parents learned that Cody had been exaggerating.}} \\
        \midrule
        \textbf{Blank Infilling} & Cody really wanted a smart phone. \\
        \midrule
        \textbf{Contextual} & Cody's parents discovered that he had been lying about wanting a smartphone when they phoned some of the other parents and found out that none of their children had ever mentioned anything about Cody wanting a new phone. \\
        \midrule
        \textbf{Prefix-Suffix} & Cody's parents phoned some of the other parents. \\
        \midrule
        \textbf{Line-by-Line} & Cody really wanted a smart phone because he thought it would make him cooler. \\
        \midrule
        \midrule
        \multicolumn{2}{C{5.25in}}{\textit{Darla was afraid of spiders. \blank She screamed and jumped away. She yelled for her father to come in and help. Her father came and removed the spider.}} \\
        \midrule
        \textbf{Blank Infilling} & Darla was afraid of spiders. \\
        \midrule
        \textbf{Contextual} & Darla was afraid of spiders because she had a bad experience with one when she was younger. \\
        \midrule
        \textbf{Prefix-Suffix} & Darla was afraid of spiders and screamed and jumped away. \\
        \midrule
        \textbf{Line-by-Line} & Darla was afraid of spiders. \\
        \bottomrule
    \end{tabular}
    \vskip -0.05in
    \caption{Zero-shot sentence infilling examples using Llama-2-7B with different prompting strategies.}
    \vskip -0.15in
    \label{tab:zero-shot-prompt-infilling}
\end{table*}
\begin{table*}[t]
    \centering
    \small
    \begin{tabular}{c | C{5.25in}} 
        \toprule
        \textbf{Method} & \textbf{Infilled Line} \\
        \midrule
        \midrule
        \multicolumn{2}{C{5.25in}}{\textit{John was at the gym. \blank He couldn't get the barbell off his chest. He was too embarrassed to call for help. John suffocated as the weights crushed his chest.}} \\
        \midrule
        \textbf{Uni-Llama} & He was on the treadmill, running at a steady pace. \\
        \midrule
        \textbf{FS-Llama} & He was doing bench presses. \\
        \midrule
        \textbf{\mbox{\sc{\textbf{MAGNET}}}} & John was doing bench presses when he accidentally dropped the barbell on top. \\
        \midrule
        \midrule
        \multicolumn{2}{C{5.25in}}{\textit{Toni took a trip to the local fair with her family. \blank The balloon was red and star shaped. Unfortunately it slipped out of her hands. She was sad to see it go.}} \\
        \midrule
        \textbf{Uni-Llama} & She won a prize for her drawing of a cat. \\
        \midrule
        \textbf{FS-Llama} & She won it in a game of darts. \\
        \midrule
        \textbf{\mbox{\sc{\textbf{MAGNET}}}} & She was so excited to see the balloon vendor and bought a red one. \\
        \midrule
        \midrule
        \multicolumn{2}{C{5.25in}}{\textit{I was resting on my couch. \blank They raised their water guns. I immediately ran. I went outside and sprayed them with the hose when they followed.}} \\
        \midrule
        \textbf{Uni-Llama} & I was feeling a bit under the weather, so I decided to take a break. \\
        \midrule
        \textbf{FS-Llama} & My neighbor's kids were playing tag in the backyard. \\
        \midrule
        \textbf{\mbox{\sc{\textbf{MAGNET}}}} & I heard a noise and looked out the window and saw two boys with water guns. \\
        \midrule
        \midrule
        \multicolumn{2}{C{5.25in}}{\textit{Kaylee owned a ferret who always ran free around the house. \blank She put flyers all over the neighborhood. There was a knock at the door three days later. Someone found her ferret.}} \\
        \midrule
        \textbf{Uni-Llama} & She loved her ferret, but she knew that ferrets have a tendency to get into things they shouldn't. \\
        \midrule
        \textbf{FS-Llama} & She put up a reward for the safe return of her ferret. \\
        \midrule
        \textbf{\mbox{\sc{\textbf{MAGNET}}}} & She was devastated when she lost him and searched everywhere for him. \\
        \midrule
        \midrule
        \multicolumn{2}{C{5.25in}}{\textit{Cody really wanted a smart phone. \blank Cody protested that all of his friends had their own phones. His parents phoned some of the other parents. His parents learned that Cody had been exaggerating.}} \\
        \midrule
        \textbf{Uni-Llama} & He had been saving up his allowance for months, but he just couldn’t. \\
        \midrule
        \textbf{FS-Llama} & Cody's friends told him they had been saving up for their phones. \\
        \midrule
        \textbf{\mbox{\sc{\textbf{MAGNET}}}} & His parents refused to buy him one because they thought it was too expensive and dangerous. \\
        \midrule
        \midrule
        \multicolumn{2}{C{5.25in}}{\textit{Darla was afraid of spiders. \blank She screamed and jumped away. She yelled for her father to come in and help. Her father came and removed the spider.}} \\
        \midrule
        \textbf{Uni-Llama} & She had always been afraid of them, even as a child. \\
        \midrule
        \textbf{FS-Llama} & She saw a spider crawling on her bed. \\
        \midrule
        \textbf{\mbox{\sc{\textbf{MAGNET}}}} & She was in her room when she saw a spider on her bed. \\
        \midrule
        \midrule
        \multicolumn{2}{C{5.25in}}{\textit{A strange structure appeared on the outside of Connie’s house one day. \blank She called an exterminator to come get them. The exterminator came and got rid of the wasps quickly. Connie was happy that the nest was gone.}} \\
        \midrule
        \textbf{Uni-Llama} & It was a large, glowing crystal that seemed to pulse with. \\
        \midrule
        \textbf{FS-Llama} & She saw a wasp flying around her kitchen window. \\
        \midrule
        \textbf{\mbox{\sc{\textbf{MAGNET}}}} & Connie noticed that there were wasps in the house and outside. \\
        \bottomrule
    \end{tabular}
    \vskip -0.05in
    \caption{Sentence infilling examples using Llama-2-7B and its adaptation using \unigr{}. Uni-Llama refers to the unidirectional model that only considers the left context, and FS-Llama is the few-shot variant that learns to use the full context to generate the infilling.}
    \vskip -0.15in
    \label{tab:infilling-examples}
\end{table*}
\begin{table*}[t]
    \centering
    \small
    \begin{tabular}{c | C{5.6in}}
        \toprule
        \textbf{Model} & \textbf{Generated Text} \\
        \midrule
        \midrule
        \multicolumn{2}{c}{Initial phrase: \highlight{The film was well received}} \\
        \midrule
        Llama & \highlight{The film was well received} by critics and audiences alike, with many praising its unique blend of humor and heart. Here are some of the key reasons why critics and audiences loved "The Big Sick":\par
        1. Unique blend of humor and heart: The film's ability to balance humor and heart was widely praised by critics and audiences. The film's lighthearted moments are often juxtaposed with more serious and emotional scenes, creating a film that is both funny and heartwarming.\par
        2. Authentic representation: The film is ... \\
        \midrule
        LLM2Vec & \highlight{The film was well received} by critics and audiences alike.\par The film was well received by critics and audiences alike.\par 
        The film was well received by critics and audiences alike.\par 
        The film was well received by critics and audiences alike.\par 
        The film was well received by critics and audiences alike.\par 
        The film was well received by critics and audiences alike.\par
        The film was well received by critics and audiences alike.\par
        The film was well received by critics and ... \\
        \midrule
        \unigr{} & \highlight{The film was well received} by critics and audiences alike and was nominated for several awards including the Academy Award for Best Visual Effects and the Saturn Award for Best Science Fiction Film. The film was also a commercial success and grossed over \$100 million at the box office.\par 
        = = Plot = = \par
        In the year 2018, a meteorite crashes into the Pacific Ocean, causing a massive tsunami that destroys most of the world's coastal cities. The survivors of the disaster band together ... \\
        \midrule
        BERT & \highlight{The film was well received} and the " " " " " " " " " " " " " " are " " are " are " are " are " are are " are are the are the the the the the the the the the the the the the so so and and and and and and and and and and and and and and and and and and and and and and and and and and and and and and and and and and and and and and and and and and and and and and ... \\
        \midrule
        \midrule
        \multicolumn{2}{c}{Initial phrase: \highlight{After 1805, the exploits of}} \\
        \midrule
        Llama & \highlight{After 1805, the exploits of} the British East India Company in India became increasingly controversial, and the company faced growing criticism from both the British public and Indian nationalists. Here are some of the key events that contributed to the decline of the British East India Company: 1. The Battle of Buxar (1757): The British East India Company defeated the French and their Indian allies in the Battle of Buxar, which gave them control over Bengal. This marked the beginning of the company's expansion in India. 2. The Battle of Plassey ... \\
        \midrule
        LLM2Vec & \highlight{After 1805, the exploits of} the French army in Egypt continued to be successful. In 1806, the French army defeated the Ottoman army at the Battle of Aboukir. In 1807, the French army defeated the Ottoman army at the Battle of Austerlitz. In 1808, the French army defeated the Ottoman army at the Battle of Wadi Abu Jarai. In 1809, the French army defeated the Ottoman army at the Battle of Roli. In 1810, the French army defeated ... \\
        \midrule
        \unigr{} & \highlight{After 1805, the exploits of} the pirates of the Caribbean became less frequent and the British Navy began to take a more active role in suppressing piracy. The British Navy began to patrol the Caribbean and the Spanish Main, and the pirates were forced to flee to other areas. The pirates of the Caribbean were not completely eliminated, however, and piracy continued to be a problem in the Caribbean until the mid 19th century. The last pirate to be executed in the Caribbean was Captain Kidd, who was hanged ... \\
        \midrule
        BERT & \highlight{After 1805, the exploits of} the the - the - way - to - the - other - to - the - other - to - the - other - to - the - other -. -. -. -. -. -. -. -. -. -. -. -. -. -. -. -. -. -. -. - " - " - " " - " " " " " " " " " " " - " " - " " " " " " " " " " and the " - " " " - " " - " - " " - " " - " " " " " " " " \\
        \bottomrule
    \end{tabular}
    \vskip -0.05in
    \caption{Text generated using greedy decoding with BERT-Base, Llama-2-7B, and its adaptations using LLM2Vec and \unigr{}. To generate text from BERT, we recursively add a mask token at the end of the input sentence and use the model to predict the mask.}
    \vskip -0.15in
    \label{tab:generation-examples}
\end{table*}

\end{document}